\documentclass[conference]{IEEEtran}
\IEEEoverridecommandlockouts
\usepackage{cite}
\usepackage{amsmath,amssymb,amsfonts}
\usepackage{algorithmic}
\usepackage{graphicx}
\usepackage{textcomp}
\usepackage{xcolor}
\def\BibTeX{{\rm B\kern-.05em{\sc i\kern-.025em b}\kern-.08em
    T\kern-.1667em\lower.7ex\hbox{E}\kern-.125emX}}
    
\begin{document}

\title{Is GPT Powerful Enough to Analyze the Emotions of Memes? \\}

\makeatletter
\newcommand{\linebreakand}{%
  \end{@IEEEauthorhalign}
  \hfill\mbox{}\par
  \mbox{}\hfill\begin{@IEEEauthorhalign}
}
\makeatother

\author{\IEEEauthorblockN{Jingjing Wang}
\IEEEauthorblockA{\textit{School of Computing} \\
\textit{Clemson University}\\
Clemson, South Carolina, USA \\
jingjiw@clemson.edu}
\and
\IEEEauthorblockN{Joshua Luo$^\ast$}
\IEEEauthorblockA{\textit{The Westminster Schools} \\
Atlanta, Georgia, USA \\
joshualuo@westminster.net}
\and
\IEEEauthorblockN{Grace Yang$^\ast$}
\IEEEauthorblockA{\textit{South Windsor High School} \\
South Windsor, Connecticut, USA \\
gy29627@southwindsorschools.org}
\linebreakand
\IEEEauthorblockN{Allen Hong$^\ast$}
\IEEEauthorblockA{\textit{D.W. Daniel High School} \\
Clemson, South Carolina, USA \\
25allenhong@pickens.k12.sc.us}
\and
\IEEEauthorblockN{Feng Luo}
\IEEEauthorblockA{\textit{School of Computing} \\
\textit{Clemson University}\\
Clemson, South Carolina, USA \\
luofeng@clemson.edu}
\thanks{$^\ast$Summer interns at the School of Computing, Clemson University.}
}

\maketitle

\begin{abstract}
Large Language Models (LLMs), representing a significant achievement in artificial intelligence (AI) research, have demonstrated their ability in a multitude of tasks. This project aims to explore the capabilities of GPT-3.5, a leading example of LLMs, in processing the sentiment analysis of Internet memes. Memes, which include both verbal and visual aspects, act as a powerful yet complex tool for expressing ideas and sentiments, demanding an understanding of societal norms and cultural contexts. Notably, the detection and moderation of hateful memes pose a significant challenge due to their implicit offensive nature. This project investigates GPT's proficiency in such subjective tasks, revealing its strengths and potential limitations. The tasks include the classification of meme sentiment, determination of humor type, and detection of implicit hate in memes. The performance evaluation, using datasets from SemEval-2020 Task 8 and Facebook hateful memes, offers a comparative understanding of GPT responses against human annotations. Despite GPT’s remarkable progress, our findings underscore the challenges faced by these models in handling subjective tasks, which are rooted in their inherent limitations including contextual understanding, interpretation of implicit meanings, and data biases. This research contributes to the broader discourse on the applicability of AI in handling complex, context-dependent tasks, and offers valuable insights for future advancements.
\end{abstract}


\section{Introduction}

Large Language Models (LLMs), representing a significant achievement in artificial intelligence (AI) research, have recently gathered substantial interest from both academia and industry. Due to their exceptional capacity to comprehend, generate, and engage in complex linguistic tasks, these models are revolutionizing the development and application of AI algorithms. Equipped with sophisticated algorithms and vast training datasets, LLMs have demonstrated advanced conversational capabilities, processing textual input and output to participate in detailed and meaningful dialogues \cite{liang2022holistic, ouyang2022training, zhao2023survey}.

 OpenAI's ChatGPT \cite{openai2023} is a prominent representative among these LLMs. It is an AI chatbot backed by the generative pretrained transformer (GPT) model and has attracted a broad societal interest. The proficiency of ChatGPT is particularly noticeable in the realm of question answering, spanning multiple sectors including, but not limited to, healthcare \cite{jeblick2022chatgpt} and finance \cite{guo2023close}. This model uses its linguistic capabilities to understand questions and generate human-like answers, marking a crucial milestone in AI-driven communication. Nevertheless, despite these impressive feats, LLMs, including GPT, confront considerable challenges when addressing subjective tasks, one such task being the interpretation and annotation of internet memes. 

\begin{figure}[ht]
  \centering
  \includegraphics[width=\linewidth]{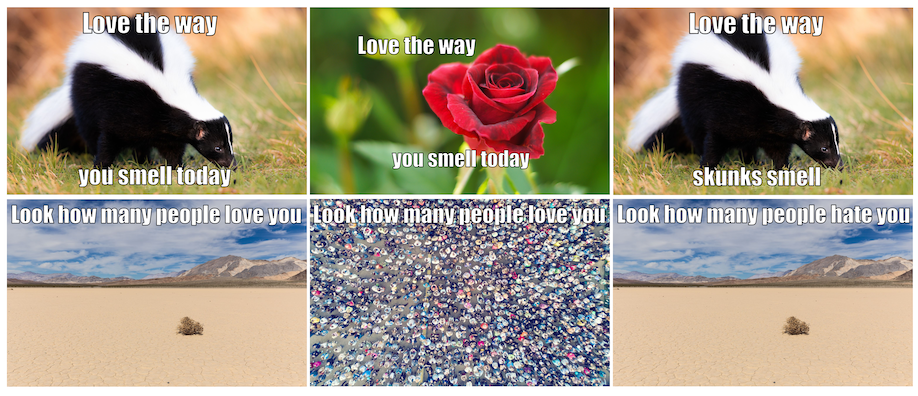}
  \caption{Hateful and non-hateful meme examples from Facebook hateful memes dataset.}
\end{figure}
Internet memes, modern digital artifacts,  encompass a variety of cultural and sociological attitude, and serve as a type of social shorthand among online communities \cite{bauckhage2011insights, shifman2013memes}.  These memes blend verbal and graphic elements to express complex concepts or feelings in a concise manner. However, the multimodal nature of memes, which combine various images with informal, often humorous or ironic text, introduces an intricate layer of complexity. To comprehend and analyze memes, one must manage the subtle interplay of visual clues, linguistic content, cultural allusions, and common community knowledge. This complexity, along with how subjective memes can be, makes it really tough for LLMs to accurately interpret memes and analyze their sentiment. 

\begin{figure}[h]
  \centering
  \includegraphics[width=\linewidth]{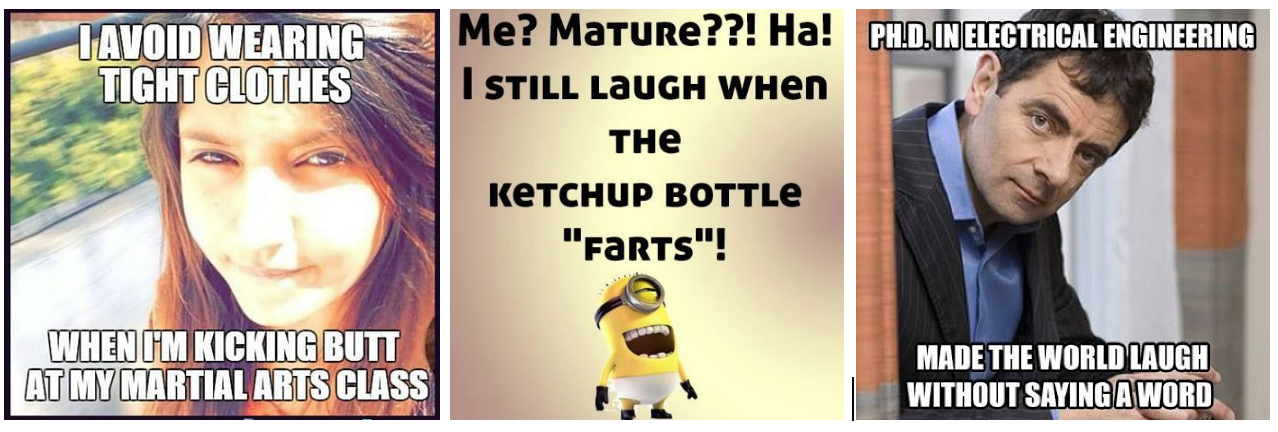}
  \caption{Meme examples from the Multimodal Memotion Analysis dataset. The score of humorous, sarcastic, and offensive from left to right is: (3,1,0),(0,3,0),(0,3,3)}
  
\end{figure}

Sentiment Analysis (SA) primarily concentrate on interpreting user sentiments from textual content. However, due to the proliferation and pervasiveness of multimodal social media data, such as online memes, SA must diversify to handle multimodal data sources. This diversification becomes even more complex when considering 'hateful memes', which cleverly combine text and images to spread offensive or harmful messages under the guise of humor \cite{kiela2020hateful, suryawanshi2020multimodal, williams2016racial}. Detecting and moderating such content not only requires understanding the text-image interaction but also a grasp of the cultural context and, in many cases, specific sub-cultural expertise. These challenges call for ongoing research and development in the field, pushing the boundaries and exploring the potential of LLMs like GPT.

In this project, we focus on the capabilities and limitations of LLMs like GPT in dealing with subjective and nuanced tasks, particularly internet meme sentiment analysis. This exploration forms the crux of this research, contributing to the broader discussion on the evolving potential and limitations of AI. In this project, we delve into the intricacies of subjectivity, assessing GPT's performance in tasks necessitating an understanding of societal norms and cultural contexts. Our investigation primarily revolves around two research questions: 1. Can GPT accurately detect implicitly hateful memes given a specific prompt? 2. Can GPT effectively conduct sentiment analysis of memes, including classifying the sentiment of each meme as positive or negative, and categorizing the type of humor into three sub-classes: sarcastic, humorous, and offensive?  We draw upon datasets from  Facebook's hateful memes \cite{kiela2020hateful} and SemEval-2020 Task 8: Multimodal Memotion Analysis \cite{sharma2020semeval} and for this project. The experimental section of the study will present a comparative analysis highlighting the constraints and limitations of GPT's responses.

\section{RELATED WORK}

\subsection{Large Language Model Prompting}

The remarkable success of LLMs, such as GPT and its variants, has also brought prompt-based learning to the fore. Prompt-based learning has found applications in various natural language processing (NLP) tasks, such as sentiment classification and natural language inference. Prompting visual-language models for computer vision tasks is another area that has started to gain attention \cite{radford2021learning, zhou2022conditional}.

However, the existing body of work in prompt-based learning has primarily focused on unimodal tasks, with limited research delving into multimodal tasks \cite{zeng2022socratic, uppal2022multimodal}. A few exceptions like the work \cite{lu2022learn} attempted multimodal tasks by utilizing the GPT-3 model \cite{brown2020language} for science question-answering tasks, where the questions have either an image or text context or both. However, these attempts have their own limitations, with models like GPT-3 being less accurate and possessing restrictive input length constraints, thereby limiting their performance and utility.

GPT-3.5-Turbo, which performs much better than GPT-3, provides an alternative solution. It presents a practical and efficient route for prompt-based learning in multimodal tasks. Therefore, our research focuses on leveraging the capabilities of the GPT-3.5-Turbo for classifying sentiments in memes, which requires the simultaneous processing of visual and textual data.

Additionally, research indicates that when equipped with a range of external NLP tools \cite{huang2022language, paranjape2023art}, Large Language Models (LLMs) can serve as effective action planners, selecting and utilizing tools for problem-solving. It suggests that these models can be extended to more complex multimodal scenarios, involving both reasoning and action, thereby enhancing their capabilities.

The GPT series has played a pivotal role in promoting prompt-based learning. We follow many of its core concepts in our work, but our focus is different. While most studies are set on applying prompting to extract knowledge from pre-trained models, our goal is to employ these techniques to fine-tune downstream tasks. Specifically, we are interested in exploring the application of these methods to the sentiment classification of memes, a complex and nuanced multimodal task.
\subsection{Hateful Meme Detection and Memotion Analysis Navigation}

Detecting hateful content in memes is a difficult process. The subtle undertones of humor and sarcasm are often interwoven within memes, coupled with the potential mismatch between the images and text, present formidable obstacles to traditional text-based hate speech detection approaches. This challenge has been the focal point of several studies, all in an effort to establish viable solutions \cite{chiu2021detecting, pramanick-etal-2021-momenta-multimodal, sharma2022detecting}.

The Hateful Memes Challenge initiated by Facebook stands as a pioneering effort in this sphere, inviting researchers worldwide to develop models capable of identifying hate speech in multimodal meme content. The dataset provided by Facebook, consisting of 10,000 manually annotated memes for hate speech, has proven to be an invaluable resource propelling research in this domain.

Mirroring these endeavors, a model that combines multi-task learning was proposed to detect hateful and offensive content present in memes. A crucial component of this model is its ability to help GPT model to view the images and do the hateful/non-hateful classification, thereby contributing to a better comprehension of hateful memes.

In spite of these extensive efforts, the task of hateful meme detection remains intricate and largely unsolved, attributed to the nuanced interaction of language and imagery inherent to the meme content. The ongoing research and continuous strides towards advancements hint at a necessity for more sophisticated models, ones that can manage these complexities and accurately pinpoint harmful or offensive content within memes.

Memotion analysis offers a captivating intersection between the realms of computer vision, natural language processing, and cognitive science. The main aim is to unpack the intricate layers of meaning, sentiment, and emotion encapsulated within memes, an endeavor that has seen many researchers strive to establish computational models to this end \cite{tanaka2022learning, scott2021memes}.

A noteworthy stride in this area was the SemEval 2020 Task 8: Multimodal Memotion Analysis \cite{sharma2020semeval}. This project tasked various models with the prediction of sentiment, humor, sarcasm and offensiveness in memes. Taking a similar direction, another researcher embarked on the creation of an automatic meme generator capable of tailoring its output based on the targeted sentiment \cite{pramanick2021momenta}.

Within this project, we proposed a similar framework for hateful meme detection tasks with some more detailed prompts by utilizing multimodal sentiment analysis and sarcasm detection to dissect visual and textual cues present in memes. The insights gathered from our work have shed light on the intricate facets of memotion analysis, showcasing the potential and limitations of AI in this field.

\section{Methodology}
\subsection{Dataset details}
 In our experimental design, we focused on two distinct datasets for sentiment analysis: the Facebook Hateful Memes dataset, and the Multimodal Memotion Analysis dataset. Each of these datasets brought unique elements to our study, enabling a multifaceted exploration of GPT's capabilities in multimodal reasoning tasks.

The Facebook Hateful Memes dataset was created with the primary intent of observing and understanding the influence of memes in propagating hate speech and offensive content. We randomly chose 200 memes from the Facebook Hateful Meme dataset for our experiments: a set of 100 randomly chosen hateful memes and the other set of 100 non-hateful memes. The primary criteria for this categorization were based on the content and contextual implications of the text interwoven within these meme images. This subset is teeming with complex emotions, often characterized by layers of sarcasm or satire. This provides an exceptionally challenging yet rewarding terrain for testing the proficiency of our approach in deciphering and correctly classifying the sentiments embedded within these memes.

The Multimodal Memotion Analysis dataset, on the other hand, is a more diverse collection of memes. We randomly selected 100 memes from this dataset, spanning a wide range of sentiments. This dataset is designed to assess the ability of the AI model to deal with a broader scope of emotions and sentiments. The memes in this dataset cover a variety of themes and concepts. They contain different types of memes, each conveying a unique sentiment. These sentiments covered range from clear-cut positive or negative overall sentiment to more complex emotions like humor, which type includes humorous, sarcastic and offensive. Furthermore, the scale is introduced to evaluate the degree of humor, where "not" corresponds to 0, "slightly" to 1, "mildly" to 2, and "very" to 3 (see Fig. 2). The broad spectrum of sentiments present in this dataset poses a significant challenge to the AI model, demanding a high degree of understanding and reasoning capability.

\subsection{Experimental Setup and Workflow}
 The main objective of the experimental setup was to configure GPT to process multimodal tasks, particularly those associated with visual content. As GPT-3.5 lacks the ability to process images, we supplemented it with the image processing capabilities of VisualGPT, which allowed GPT to 'see' the images and generate a description of each image's visual content.

To guide the model's reasoning process, we used the TaskMatrix template, a tool developed by Microsoft. The TaskMatrix enabled us to construct a sequence of tasks for GPT that required it to draw upon and exercise its multimodal reasoning skills.

\begin{figure}[htbp]
\centerline{\includegraphics[width=\linewidth]{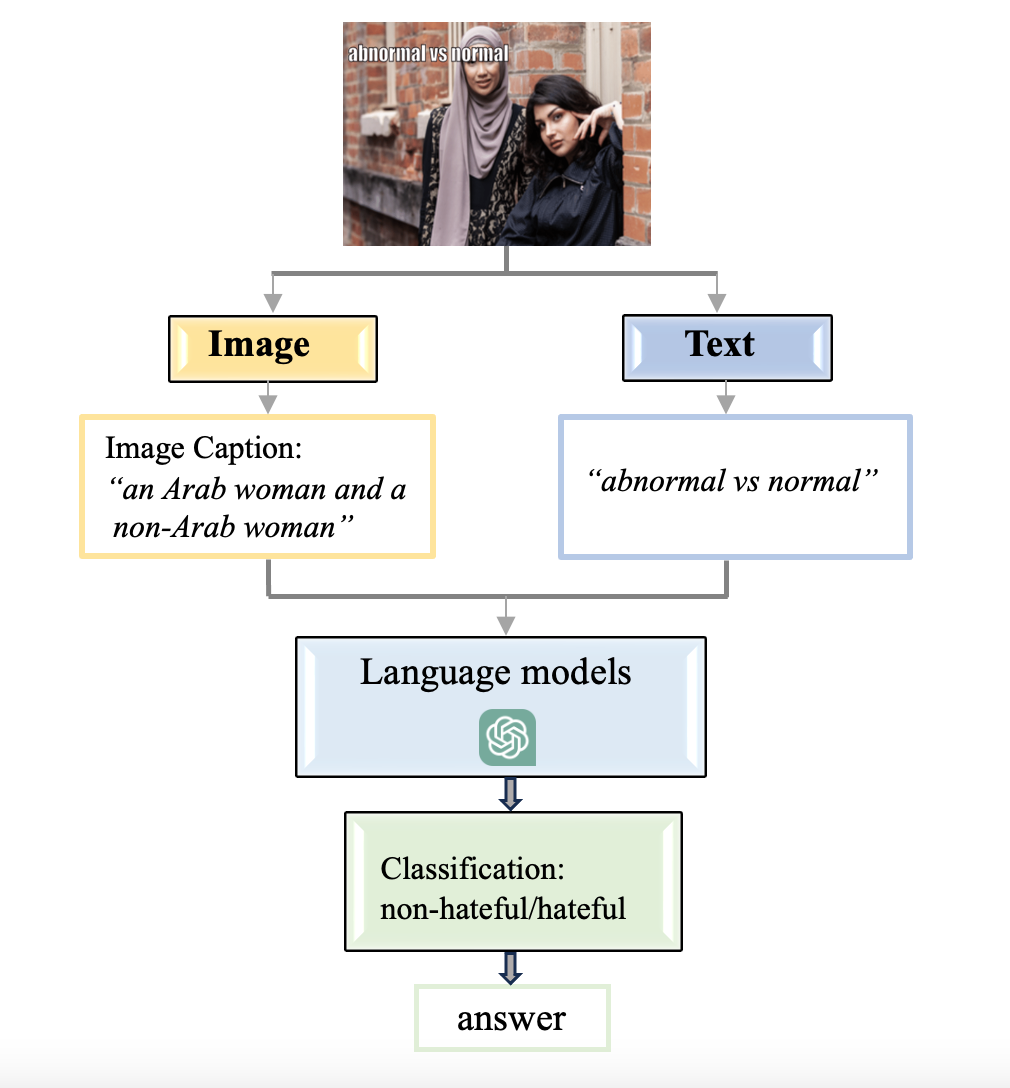}}
\caption{The workflow of Facebook hateful memes detection. "Classify the sentiment of the previous image and its accompanying text as either hate or non-hate. Use a maximum of 1 word in your response. Note that the accompanying text is: {accompanying text}."}

\end{figure}

The experimental workflow was divided into several phases: 

Image Processing Phase: The image processing phase was the initial stage of the procedure. The image analysis capabilities of VisualGPT were used to analyze the images from the datasets during this phase. VisualGPT was used to build a descriptive representation of the picture's content for each image. The description was then fed into GPT, allowing it to 'see' the image. This procedure effectively endowed GPT with the capability to 'understand' the visual content in the images, an essential prerequisite for performing sentiment analysis on these images; 

Sentiment Analysis Phase: Upon receiving the image descriptions from VisualGPT, the next stage was to utilize these descriptions to conduct sentiment analysis. For the Facebook hateful memes dataset, we employed a pre-defined notion of 'hateful sentiment'. We tasked GPT with determining if the sentiment associated with each image, as conveyed through the image's description, was hateful or non-hateful (see Fig. 3 as illustration). For the Google memotion dataset, GPT was asked to provide a generalized assessment of whether the overall sentiment of the image was positive or negative. In addition, we asked GPT to give us a detailed rating of each image in certain categories like humor, sarcasm, and offensiveness. The model was directed to rate the images in these categories on a scale from 0 to 3 (see Fig. 4 as illustration); 

Data Storage Phase: Once the model's responses were obtained, they were collected and organized in the data storage phase. The results corresponding to each image, including the image's original sentiment label and GPT's sentiment assessment, were recorded and stored in a csv file. This file was then used for the subsequent analysis of the model's performance; 

Accuracy Analysis Phase: The final phase of the experimental workflow was the accuracy analysis phase. In this phase, a specially developed program compared GPT's sentiment assessments with the original sentiment labels of the images to evaluate the model's accuracy. The accuracy was quantified as the proportion of images for which GPT's sentiment assessment matched the original sentiment label.

\begin{figure}[htbp]
\centerline{\includegraphics[width=\linewidth]{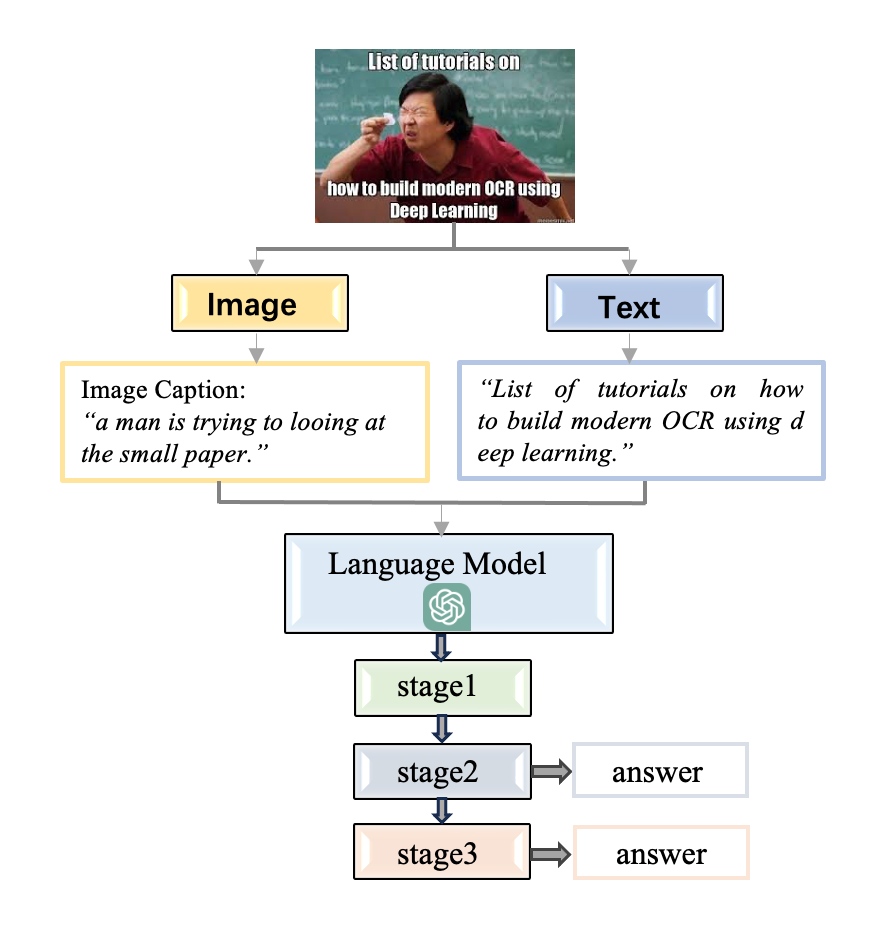}}
 \caption{The workflow of Multimodal Memotion Analysis and an example of the prompt. Stage 1:
Do not respond to this prompt. Just note that the text accompanying the previous meme is and use this information in future queries: "+text
Stage 2:
Using no more than 2 words describe the overall sentiment of the previous meme as either positive or negative. Assume the meme is not neutral and must be either positive or negative. Provide only a classification label using only 'Positive' or 'Negative' (use no more than 2 words)
Stage 3:
On a scale of 0 to 3 quantify the previous meme in all of the following categories: humour, sarcastic and offensive. Do not provide anything except the classification and degree. Answer with only the label and then the degree in this format: humorous(x) sarcastic(x) offensive(x)  .}

\end{figure}

\section{Experiment Results}

In this section, we conducted an in-depth investigation by using the two datasets aforementioned in section 3 to explore the effectiveness of GPT in analyzing sentiments in memes. After carefully evaluating each meme based on the prompts designed, we meticulously collated and analyzed the results. Here is the comprehensive analysis of our findings.

{\bfseries Case 1: The results of Facebook Hateful Memes dataset}

Our initial investigation involved the evaluation of GPT's proficiency in identifying hateful content within memes using Facebook's Hate/Non-Hate Classification Dataset. Here we repeat the experiments four times with different prompts.

Hateful Content Identification: The task of identifying hateful memes posed a significant challenge. It required the AI to understand not only the content of the image and the accompanying text but also their nuanced interplay. The results showed an accuracy rate of 39\% in detecting hateful content with a relatively high standard deviation of 6.87\%. It can be seen that the performance of GPT is far less accurate in classifying the hateful memes, and the different prompts would affect the decision of GPT considerably. To accurately flag hateful memes, a model needs to be adept at grasping not only the overt message but also the undertones, context, and subtext. The 39\% accuracy rate thus provides us with a baseline for improvement and future model refinements.

Non-Hateful Content Identification: On a brighter note, GPT's performance in recognizing non-hateful memes was much stronger, with an impressive accuracy rate of 80\% and much lower standard deviation of 2.69\%. The high accuracy suggests that GPT is quite adept at discerning memes that are not hateful, and the different prompts used had minimal impact on its decision-making process.

\begin{table}[htbp]
  \caption{The Result of Facebook Hateful Memes dataset}
  \label{tab:freq}
  \centering
  \begin{tabular}{|c|c|}
    \hline
    Model & \multicolumn{1}{|c|}{VisualGPT+GPT-3.5-Turbo} \\
    \hline
    Hateful Memes Accu (sd) & 39\% (6.87\%) \\
    \hline
    Non-Hateful Memes Accu (sd) & 80\% (2.69\%) \\
    \hline
  \end{tabular}
\end{table}

{\bfseries Case 2: Memotion analysis dataset}

Overall Sentiment Classification: the first task in this series was classifying the overall sentiment of a meme as positive, negative, or neutral. In this regard, GPT achieved an accuracy rate of 79\% for positive and 35\% for negative. It can be observed that this result is consistent with the performance in the classification of hateful memes in the Facebook dataset, which shows confidence in understanding the positive sentiment in a meme, based on both the visual content and the accompanying text. For negative (for example, offensive in the later paragraph) content detection, the low accuracy still suggests the challenge.

Humor recognition: GPT achieved a 60\% accuracy rate in the task of recognizing humor. This outcome is notably remarkable, considering the intricate nature of humor as a multifaceted human emotion, often subject to cultural, social, and individual perspectives. The considerable level of accuracy exhibited by GPT indicates its ability to comprehend the humorous  aspects present in a diverse range of memes. This result also implies that the GPT has acquired the ability to recognize various indicators of humor to a certain degree. These indicators might be obvious, such as punchlines or visual gags, or implicit, such as irony or absurdity.

Sarcasm recognition: the model's performance in the domain of sarcasm recognition was notably lower, registering an accuracy of 45\% as illustrated in Table 2. Though lower, this accuracy level is acceptable given the inherent difficulties of sarcasm detection. Sarcasm often involves expressing the opposite of what is meant, requiring an understanding of context, tone, and frequently shared cultural knowledge. 
This observation helps to emphasize the limits of the approach in accurately detecting more subtle emotional cues and contextual subtleties. The lower accuracy rate observed in sarcasm recognition highlights the need for further study and model enhancement in this area.

Offensive recognition: as seen in Table 2, GPT's accuracy stood at 37\% in recognizing offensive content, presenting the weakest performance. As aforementioned in overall sentiment classification, GPT meets the huge challenge in offensive content recognition. The consistency of this finding illustrates the difficulty associated with detecting offensive content.  
Comparable to hateful meme classification, offensive content in memes may exhibit subtlety, sometimes concealed within humor or sarcasm, or manifest through a complex interplay of textual and visual elements that can be hard to interpret for an AI.

\begin{table}[htbp]
  \caption{Accuracies of Different Emotions}
  \label{tab:model-accuracy}
  \centering
  \begin{tabular}{|c|c|}
    \hline
    Model & \multicolumn{1}{|c|}{VisualGPT+GPT-3.5-Turbo} \\
    \hline
    Humor Scale Accu & 60\% \\
    \hline
    Sarcasm Scale Accu & 45\% \\
    \hline
    Offensive Scale Accu & 37\% \\
    \hline
    Overall Accu (Positive) & 79\% \\
    \hline
    Overall Accu (Negative) & 35\% \\
    \hline
  \end{tabular}
\end{table}

\section{Discussion}
In this section, we discuss the limitations of GPT model, the conclusion of our study and future work.

\subsection{Limitations}
From the experiment results, we can learn that the current GPT-3.5 model still has some limitations.

Lack of Deep Contextual Understanding: GPT model, despite its advanced capabilities, is still lacking a deep understanding of human nuances, societal norms, and cultural contexts, which are often critical to accurately interpreting and responding to subjective tasks. This limitation becomes particularly pronounced when the GPT model is faced with local idioms, colloquialisms, or culturally specific references.

Difficulty in Understanding Implicit Meaning: LLMs typically struggle with interpreting and generating content that contains implicit or hidden meanings, especially those requiring an understanding of human emotions, intentions, or sarcasm. This limitation becomes particularly apparent in tasks involving the interpretation of sarcasm or offensiveness in memes, as these tasks often require a nuanced understanding of cultural or social contexts.

Biases in Training Data: LLMs are trained on vast amounts of data from the internet, which may contain various forms of biases. The presence of biases inside the model might accidentally impact its responses in subjective tasks, leading to much less accurate outcomes. 

\subsection{Conclusion and Future Work}
Our efforts to assess how well GPT could analyze sentiments in memes led us to some interesting and insightful findings. 

On one hand, the model performed impressively when it came to classifying non-hateful memes and understanding the positive mood of a meme. The significant accuracy achieved in non-hateful meme classification, positive sentiment determination, and humor recognition, demonstrate the GPT's ability to comprehend and interpret the content in a majority of memes correctly.

However, the accuracy rate dropped a lot when encountered with detection of hateful, sarcasm and offensive content in memes. The lower accuracy rates seen in these categories shows the intricate nature of identifying concealed hateful or offensive content inside memes.

As GPT-4 released in recent days, it will be beneficial if we can integrate the latest model in our framework; moreover, as fine-tuning GPT-3.5-Turbo is available, we can fine tune our exist models to make some improvements in its ability to detect hateful and offensive content. We leave these as future work.


\vspace{12pt}

\end{document}